\documentclass[10pt,twocolumn,letterpaper]{article}

\usepackage{cvpr}              
\definecolor{cvprblue}{rgb}{0.21,0.49,0.74}
\usepackage[pagebackref,breaklinks,colorlinks,allcolors=cvprblue]{hyperref}

\usepackage{url}
\usepackage{booktabs}       
\usepackage{amsfonts}       
\usepackage{nicefrac}       
\usepackage{microtype}      
\usepackage[table]{xcolor}         

\usepackage{amsmath}
\usepackage{graphicx}
\usepackage{enumitem}
\usepackage{makecell}
\usepackage{multirow}
\usepackage{arydshln}
\usepackage{subcaption}
\usepackage{amssymb}
\usepackage{pifont}
\usepackage[accsupp]{axessibility}  
\usepackage{accents}
\usepackage{algorithm}
\usepackage{algpseudocode}

\usepackage{color,soul}

\definecolor{first}{RGB}{255, 172, 172}   
\definecolor{second}{RGB}{255, 208, 142}  
\definecolor{third}{RGB}{255, 255, 173}   

\definecolor{chosen}{RGB}{180,255,180}  

\usepackage{array}
\newcolumntype{$}{>{\global\let\currentrowstyle\relax}}
\newcolumntype{^}{>{\currentrowstyle}}
\newcommand{\rowstyle}[1]{\gdef\currentrowstyle{#1}%
  #1\ignorespaces
}

\newcommand{\methodname}{Long-LRM++}
\newcommand{\projectpage}{\url{http://arthurhero.github.io/projects/llrm2/}}

\def\paperID{****} 
\def\confName{CVPR}
\def\confYear{2026}

\title{Long-LRM++: Preserving Fine Details in Feed-Forward Wide-Coverage Reconstruction}

\author{
Chen Ziwen$^1$~~~
Hao Tan$^1$~~~ 
Peng Wang$^2$~~~
Zexiang Xu$^3$~~~
Li Fuxin$^4$~~~\\
\vspace{-0.1in}
\normalsize
$^1$Adobe Research~~~
$^2$Tripo AI~~~
$^3$Hillbot~~~
$^4$Oregon State University\\
}

\begin{document}
\maketitle
\begin{abstract}

Recent advances in generalizable Gaussian splatting (GS) have enabled feed-forward reconstruction of scenes from tens of input views. Long-LRM notably scales this paradigm to 32 input images at 960×540 resolution, achieving 360° scene-level reconstruction in a single forward pass. However, directly predicting millions of Gaussian parameters at once remains highly error-sensitive: small inaccuracies in positions or other attributes lead to noticeable blurring, particularly in fine structures such as text. 
In parallel, implicit representation methods such as LVSM and LaCT have demonstrated significantly higher rendering fidelity by compressing scene information into model weights rather than explicit Gaussians, and decoding RGB frames using the full transformer or TTT backbone. 
However, this computationally intensive decompression process for every rendered frame makes real-time rendering infeasible.
These observations raise key questions: 
Is the deep, sequential ``decompression” process necessary? Can we retain the benefits of implicit representations while enabling real-time performance?
We address these questions with Long-LRM++, a model that adopts a semi-explicit scene representation combined with a lightweight decoder. Long-LRM++ matches the rendering quality of LaCT on DL3DV while achieving real-time 14
FPS rendering on an A100 GPU, overcoming the speed limitations of prior implicit methods. Our design also scales to 64 input views at the 950×540 resolution, demonstrating strong generalization to increased input lengths. 
Additionally, Long-LRM++ delivers superior novel-view depth prediction on ScanNetv2 compared to direct depth rendering from Gaussians. Extensive ablation studies validate the effectiveness of each component in the proposed framework.
Project page: \projectpage

\end{abstract}    
\vspace{-0.2in}

\section{Introduction}
\label{sec:intro}

\begin{figure}[h]
    \centering
    \includegraphics[width=\linewidth]{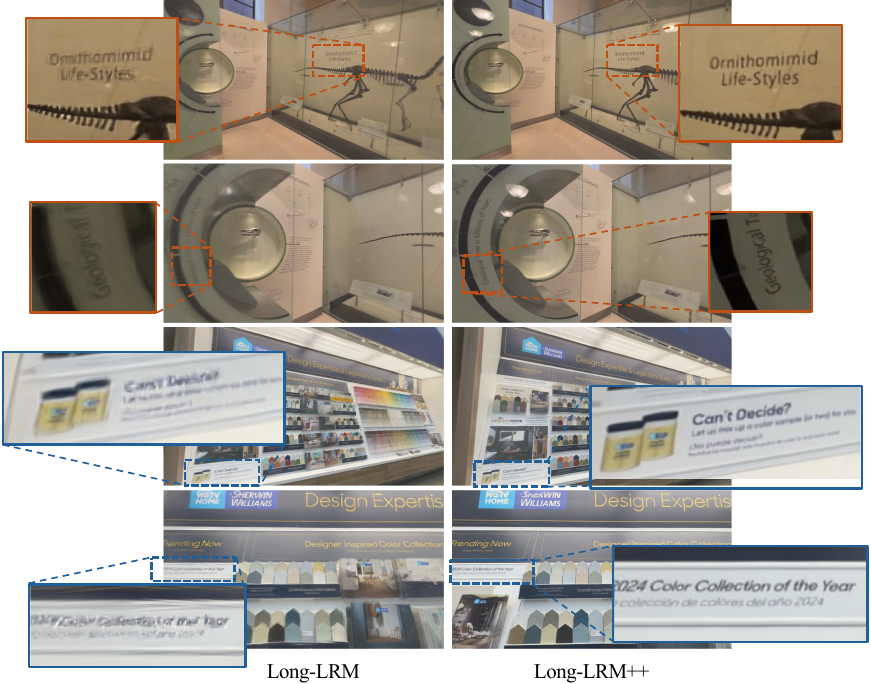}
    \vskip -0.05in
    \caption{\small 
    We present \methodname{}, a feed-forward novel-view synthesis method for high-resolution scene reconstruction. By leveraging a semi-explicit feature-Gaussian representation, \methodname{} substantially reduces the blurriness issues observed in Long-LRM while preserving real-time rendering speed.
}
\vspace{-0.2in}
    \label{fig:teaser}
\end{figure}

In recent years, novel-view synthesis has emerged as a prominent approach for 3D reconstruction, enabling the generation of photorealistic renderings of real-world scenes from arbitrary viewpoints. The milestone work NeRF~\cite{nerf} encodes scene information implicitly in the weights of a multilayer perceptron (MLP), which predicts color and density for any 3D location given a viewing direction. While this implicit formulation offers a compact representation, NeRF requires hours of optimization per scene and remains far from real-time during rendering.
In contrast, 3D Gaussian Splatting (3D GS)~\cite{3dgs} introduces a fully explicit representation using a set of 3D Gaussians parameterized by position, color, and opacity. This explicit formulation enables efficient optimization—typically within minutes per scene—and achieves real-time rendering without any neural network inference.


Following the success of 3D GS, several generalizable Gaussian Splatting (GS) methods~\cite{pixelsplat,mvsgaussian,mvsplat,depthsplat,gslrm,llrm} have been developed. By training on large-scale multi-view datasets, these approaches enable instant, feed-forward GS-based reconstruction. Among them, Long-LRM~\cite{llrm} achieves, for the first time, wide-coverage scene-level reconstruction at 950×540 resolution. However, its reconstructions often appear blurry, particularly in regions with fine textures or high-frequency details such as text.
In parallel, implicit representation–based feed-forward methods~\cite{lvsm,lact} have emerged with better rendering quality. LVSM~\cite{lvsm} (\textit{decoder-only}) feeds the source images together with the target camera embedding as inputs to its pure transformer backbone.
LaCT~\cite{lact} further introduces a large-chunk test-time training layer, encoding reconstruction information in the ``fast weights", successfully scaling to large input sizes comparable to Long-LRM while delivering superior rendering quality.
Nevertheless, these implicit representation methods sacrifice a key advantage of GS-based approaches—real-time rendering, because each target frame rendering requires a full (often 24-layer) network forward pass. This trade-off is inherent: the more implicitly scene information is stored, the longer and more complex the decoding process must be.


Is it possible to overcome this trade-off and achieve better reconstruction while maintaining real-time rendering speed?
We argue that a key limitation of feed-forward GS methods lies in the \textit{color splatting} mechanism. Since each Gaussian represents only a single color, faithfully reproducing fine details in a scene requires predicting tens of millions of Gaussians—often one per pixel—to achieve high-resolution rendering. This imposes a heavy burden on feed-forward models that must predict all of the Gaussian parameters instantaneously. Moreover, even slight inaccuracies in the input camera poses or Gaussian positions can propagate and manifest as noticeable blurriness in the rendered results.


 
With this observation, we propose \methodname{}, which leverages a \textit{semi-explicit} representation to enhance Gaussian splatting for scene reconstruction. Similar to standard 3D GS, our approach maintains a set of 3D Gaussians positioned in an explicit 3D space with real-world metric correspondence. However, unlike conventional Gaussian splatting, \methodname{} relaxes the strict alignment between Gaussian positions/colors and the actual scene surfaces.
In \methodname{}, each Gaussian carries a feature vector rather than a fixed color, similar in spirit to neural point clouds. Unlike prior neural point cloud–based NVS methods~\cite{neuralpoint,pointnerf}, which initialize points using depth sensors or off-the-shelf multi-view stereo algorithms, the Gaussians in \methodname{} are free to distribute anywhere in 3D space.
During rendering, we first render a feature map at the target view using standard Gaussian splatting. This feature map is then decoded by a lightweight five-layer decoder into RGB color or depth. A key component of the decoder is the \textbf{translation-invariant local-attention block}, which significantly improves rendering quality compared to global attention with absolute positional embeddings. We further introduce a \textbf{multi-space partitioning step} that enables independent rendering and decoding of mutually exclusive Gaussian subsets, followed by feature merging before the final output layer. This design again substantially improves rendering quality.


\methodname{} combines the best of both worlds: it achieves rendering quality boost similar to LaCT while maintaining real-time rendering speed. We evaluate \methodname{} on two tasks: novel-view color rendering on DL3DV~\cite{dl3dv} at 960×540 resolution with up to 64 input views, and novel-view color and depth rendering on ScanNetv2~\cite{scannet} at 448×336 resolution with up to 128 input views. In both settings, \methodname{} achieves state-of-the-art results in terms of quality and efficiency. Its rendering quality significantly surpasses Long-LRM, despite using only 1/4 of its Gaussians. Compared to LaCT, \methodname{} achieves comparable rendering fidelity while being substantially faster in both reconstruction and rendering, maintaining a real-time 14 FPS rendering speed on a single A100 GPU.
We further conduct extensive ablation studies on key design choices, including feature Gaussian prediction and rendering, the target-frame decoder architecture, and the multi-space partitioning design. In summary, \methodname{} demonstrates that feed-forward novel-view synthesis models can achieve both quality and efficiency for high-resolution, wide-coverage, scene-level reconstruction.

\begin{figure*}[h]
    \centering
    \includegraphics[width=\linewidth]{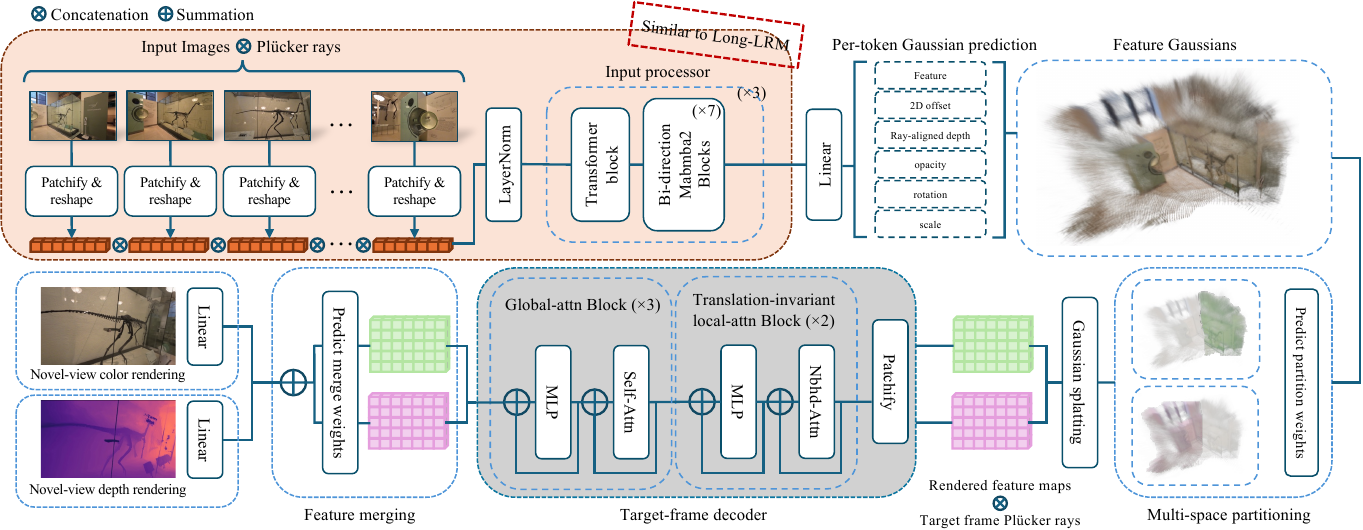}
    \vskip -0.05in
    \caption{\small 
    Overview of the \methodname{} architecture. \methodname{} takes up to 64 input images at 950×540 resolution along with their camera poses, and processes them using a backbone composed of interleaved Mamba2 and Transformer blocks, similar to Long-LRM. Each image token predicts $K$ free-moving feature Gaussians (visualized with originating pixel's color). During rendering, we introduce a multi-space partitioning step that divides the Gaussians into multiple subsets, each rendered and decoded independently. The target-frame decoder incorporates translation-invariant local-attention blocks to improve robustness and rendering quality. Finally, the decoded feature maps are merged and passed through a linear layer to produce the novel-view color or depth rendering.
}
    \label{fig:archi}
    \vspace{-0.2in}
\end{figure*}

\section{Related Work}
\label{sec:related}

\noindent\textbf{Reconstruction representations in novel-view synthesis.}
A wide range of scene representations have been explored for novel-view synthesis, spanning different levels of explicitness to balance rendering quality and efficiency. Before NeRF~\cite{nerf}, there have been methods employing neural point clouds~\cite{neuralpoint}, neural textures on meshes~\cite{deferredneuralrender}, neural volumes~\cite{stableview,deepvoxels,neuralvolume}, multiplane images~\cite{llff}, and MLP-based implicit fields~\cite{srn}. Neural Point-Based Graphics~\cite{neuralpoint}, for instance, initializes a point cloud from depth sensors, associates each point with a learnable descriptor, and renders novel views by decoding the projected features with a U-Net.
Building upon NeRF, NSVF~\cite{nsvf} learns a sparse voxel octree via differentiable ray marching, Plenoxels~\cite{plenoxels} optimizes a sparse 3D grid with spherical harmonics, Instant-NGP~\cite{instantngp} accelerates NeRF with multiresolution hash encodings, TensoRF~\cite{chen2022tensorf} factorizes the scene into low-rank tensors, and Point-NeRF~\cite{pointnerf} leverages neural point clouds initialized from off-the-shelf MVS methods.
Among Gaussian-based methods, Scaffold-GS~\cite{scaffoldgs} predicts anchor Gaussians with learnable offsets, while SAGS~\cite{sags} applies graph neural networks over Gaussian neighborhoods. Octree-GS~\cite{octreegs} introduces octree-structured anchors to enable large-scale reconstruction. All these approaches, however, rely on per-scene optimization rather than feed-forward inference.


\noindent\textbf{Feed-forward novel-view synthesis.}
Both NeRF and 3D Gaussian Splatting (3D GS) have inspired feed-forward variants trained on large-scale datasets for instant reconstruction. Both NeRF-based approaches~\cite{pixelnerf,ibrnet,neuray,gprn,mvsnerf,geonerf,enerf,zhang2022nerfusion,hong2024lrm,wang2023pflrm} and Gaussian-based methods~\cite{pixelsplat,mvsplat,mvsgaussian,depthsplat,gslrm,llrm,longsplat} typically involve volume rendering in their inference step. Among them, Long-LRM~\cite{llrm} is the first to achieve high-resolution, wide-coverage scene-level reconstruction.
In contrast, recent implicit representation methods~\cite{lvsm,lact} bypass volume rendering and directly predict RGB colors. LVSM~\cite{lvsm} employs a pure transformer backbone, achieving much higher fidelity than prior feed-forward models, but is still limited to low resolution and few input views. LaCT~\cite{lact} scales to dozens of inputs by storing scene information in the fast weights of TTT layers, yielding substantially better rendering quality than Long-LRM. However, this implicit representation requires running the entire 24-layer backbone to render each frame, significantly limiting rendering speed.
\methodname{} gets the best of both worlds by retaining the fast, explicit nature of 3D Gaussians while replacing the error-prone color splatting with the more robust feature splatting, coupled with a novel, lightweight yet powerful decoder.


\noindent \textbf{Feature rendering with 3D Gaussians.} Several works also extend 3D Gaussians with feature vectors and render feature maps, but their goals differ from ours. Feature 3DGS~\cite{feature3dgs} and Feature Splatting~\cite{featuresplat} augment each Gaussian with semantic features extracted from pre-trained 2D vision-language models such as CLIP~\cite{clip}, enabling downstream tasks like novel-view segmentation and scene decomposition. Spacetime Gaussian Feature Splatting~\cite{spacetimefeaturesplat} is an optimization-based 4D reconstruction method that replaces spherical harmonics with a 9-dimensional feature vector encoding RGB, view direction, and time, to more compactly represent radiance. BBSplat~\cite{bbsplat} and its feed-forward extension LGTM~\cite{lgtm} print color texture maps on 2D Gaussians.

\section{Method}
\label{sec:method}

\methodname{} is a feed-forward novel-view synthesis model that takes a set of RGB input images and produces the RGB color or depth map at a target camera pose. The overall architecture consists of three main components: an input processing backbone, feature Gaussian prediction and rendering, and a target-frame decoder. We describe the input processing backbone in Sec.~\ref{sec:processor}, the  feature Gaussians prediction and rendering in Sec.~\ref{sec:gs_pred}, the target-frame decoding in Sec.~\ref{sec:decoder}, and the training objectives in Sec.~\ref{sec:loss}.


\subsection{Input Processing Backbone}\label{sec:processor}

\methodname{} adopts an input processing backbone similar to Long-LRM~\cite{llrm}. The input RGB images $\mathbf{I}^{\text{input}}\in [0,1]^{V\times H\times W\times 3}$ are first patchified through a linear projection into features of shape $\mathbb{R}^{V\times \frac{H}{p}\times \frac{W}{p}\times D}$, and then reshaped into a row-major token sequence $\mathbb{R}^{L\times D}$, where $L = V \times \frac{H}{p} \times \frac{W}{p}$. The tokens are processed by a backbone composed of a mixture of Mamba2 and transformer blocks, following the design of~\cite{llrm}, which has demonstrated better scalability than pure transformer backbones and higher rendering quality than pure Mamba-based ones.
Different from~\cite{llrm}, we employ a \{1T7M\}$\!\times\!3$ block sequence, reduce the hidden dimension $D$ to 768, and remove the token merging module. Since \methodname{} uses a semi-explicit scene representation—feature Gaussians, it no longer requires dense, pixel-aligned Gaussians to capture high-frequency details for high-quality rendering, and thus only needs 1/4 number of Gaussians compared to Long-LRM, which significantly lowers GPU memory usage. The reduced memory footprint allows us to train at our highest-resolution setting without the token merging module, which, although computationally efficient, slightly degrades performance. 


\subsection{Feature Gaussian prediction and splatting}\label{sec:gs_pred}

Instead of predicting one Gaussian per pixel as in~\cite{llrm}, \methodname{} predicts $K$ Gaussians per token through a linear projection.  Each Gaussian is associated with a feature vector $\mathbf{f}\in\mathbb{R}^{F}$, a 2D offset $\mathbf{o}\in\mathbb{R}^{2}$ from the patch center, a camera-ray–aligned depth, and other Gaussian attributes including rotation, opacity, and scale. The 2D offsets allow the Gaussians to move freely from the patch centers, providing great spatial flexibility. The Gaussians' 3D positions can be calculated combining the 2D offsets and the ray-aligned depths.
Given the predicted feature Gaussians and a target camera pose, we splat the Gaussian features onto a canvas of size $\frac{H}{r}\!\times\!\frac{W}{r}$, where $H\!\times\!W$ denotes the input image resolution and $r$ is a hyperparameter. Different from color rendering with spherical harmonics, the feature vectors are view-independent. To incorporate view-dependent information, we compute the Plücker rays of the target camera and concatenate them with the rendered feature map, resulting in a final tensor of shape $\mathbb{R}^{\frac{H}{r}\times \frac{W}{r}\times (F+6)}$.

\subsection{Target-frame decoder}\label{sec:decoder}

Similar to the input images, the rendered feature map is patchified through a linear layer into shape $\mathbb{R}^{\frac{H}{q}\times \frac{W}{q}\times E}$, where $E$ denotes the hidden dimensionality of the decoder. The decoder is composed of several transformer-based attention blocks. If we were to employ only global self-attention, the same object rendered in different spatial regions would receive distinct positional embeddings, leading to inconsistent computation. To address this, the initial layers of the decoder adopt translation-invariant local-attention blocks, ensuring that the same object is processed consistently regardless of its location on the feature map. These local-attention layers are followed by several global-attention blocks, which aggregate global context across the entire frame.
We provide an ablation study of the decoder design in Table~\ref{tb:decoder}. Finally, the decoded feature map is passed through a linear projection head to produce the final RGB color or depth output.

\noindent\textbf{Translation-invariant local-attention block.}
In the local-attention layer, each token attends only to its $k$ nearest neighbors, gathering features locally. Instead of using global absolute positional embeddings, we adopt \textit{relative} positional embeddings. Let $\mathbf{\Delta x} = (\Delta x, \Delta y)$ denote the relative position between two tokens. Following~\cite{autofocusformer}, we compute
\begin{equation}
    \mathbf{r} = (\Delta x,\Delta y, \|\mathbf{\Delta}\mathbf{x}\|_2,\frac{\Delta x}{\|\mathbf{\Delta}\mathbf{x}\|_2}, \frac{\Delta y}{\|\mathbf{\Delta}\mathbf{x}\|_2})
\end{equation}
where the Euclidean distance $\|\mathbf{\Delta}\mathbf{x}\|_2$ is rotation-invariant, and the cosine and sine values are scale-invariant. Then,  $\mathbf{r}$ is passed through a small MLP and the resulting embeddings are added to the $QK$ attention scores before the softmax. In our experiments, we set the neighborhood size to $k=64$.

\noindent\textbf{Multi-space partitioning and merging.}
Inspired by~\cite{multispace}, we propose a Gaussian partitioning step that divides the set of feature Gaussians into multiple subsets, which are rendered and processed by the decoder independently, and merged before the final linear layer. One motivating example for this design is mirrors: by rendering the virtual scene inside a mirror independently of the real world and performing adaptively weighted merging, the virtual Gaussians need not adhere to the physical geometry of the real world, e.g., no virtual objects appear when viewing the back of the mirror.
The partitioning and merging work simply. Each Gaussian’s feature vector is passed through a linear layer to produce $S$ softmax-normalized weights. The Gaussians are then duplicated into $S$ copies, with each copy’s opacity values suppressed by the corresponding weights. The $S$ Gaussian splits are then rendered and decoded independently. The resulting $S$ feature maps are passed through another linear layer to produce $S$ maps of merging weights. The final feature map is obtained via a weighted-sum operator and is fed into the output linear layer to produce RGB color or depth. Empirically, $S=2$ provides a substantial performance boost compared to $S=1$ (no partitioning) with minimal impact on rendering speed. Ablation results are reported in Table~\ref{tb:featsplit}.

%

\subsection{Training objectives}\label{sec:loss}

To supervise RGB color rendering at novel camera poses, \methodname{} adopts the same training objectives as~\cite{gslrm, llrm}, combining Mean Squared Error (MSE) loss and Perceptual loss:
\begin{equation}
\footnotesize
    \mathcal{L}_\text{color}= \text{MSE}\left(\mathbf{I}^{\text{gt}}, \mathbf{I}^{\text{pred}}\right) + \lambda \cdot \text{Perceptual}\left(\mathbf{I}^{\text{gt}}, \mathbf{I}^{\text{pred}}\right)
\end{equation}
where $\lambda$ is empirically set to 0.5. For depth rendering at novel camera poses, \methodname{} adopts the log-L1 loss
\begin{equation}
\footnotesize
    \mathcal{L}_\text{depth}= \text{Smooth\_L1}\left(\log\mathbf{D}^{\text{gt}}, \log\mathbf{D}^{\text{pred}}\right)
\end{equation}
. To further improve depth map training, we adopt two auxiliary losses from~\cite{simplerecon} that enforce consistency between order-1 gradients and normals derived from the depth maps. For both losses, the predicted and ground-truth depth maps are first downsampled to 1/4 of the original resolution, yielding $\mathbf{D}_{4\downarrow}^{\text{pred}}$ and $\mathbf{D}_{4\downarrow}^{\text{gt}}$. The gradient loss is then computed as
\begin{equation}
\footnotesize
    \mathcal{L}_\text{grad}= \sum_{l=1}^4
\text{L1}\left(\text{Sobel\_filter}\left(\mathbf{D}_{4l\downarrow}^{\text{gt}}\right), \text{Sobel\_filter}\left(\mathbf{D}_{4l\downarrow}^{\text{pred}}\right)\right)
\end{equation}
where the Sobel filter is applied along both vertical and horizontal directions. The normal loss is computed by first unprojecting the downsampled depth maps into 3D point maps, $\mathbf{P}_{4\downarrow}^{\text{pred}}$ and $\mathbf{P}_{4\downarrow}^{\text{gt}}$. Sobel filters are then applied in both directions to approximate the normal maps, $\mathbf{N}_{4\downarrow}^{\text{pred}}$ and $\mathbf{N}_{4\downarrow}^{\text{gt}}$. Finally, a cosine similarity loss is applied:
\begin{equation}
\footnotesize
    \mathcal{L}_\text{normal}= 1-
\text{Cosine\_similarity}\left(\mathbf{N}_{4\downarrow}^{\text{gt}}, \mathbf{N}_{4\downarrow}^{\text{pred}}\right)
\end{equation}.
The ablation results for the two auxiliary losses are reported in Table~\ref{tb:dep_ab}. The overall training objective for depth rendering is then defined as:
\begin{equation}
    \footnotesize
    \mathcal{L}_\text{depth\_total}= \mathcal{L}_\text{depth}+\lambda_{\text{grad}}\cdot \mathcal{L}_\text{grad}+\lambda_{\text{normal}}\cdot \mathcal{L}_\text{normal}
\end{equation}
where we empirically set $\lambda_{\text{grad}}=\lambda_{\text{normal}}=0.1$.

We omit the opacity loss from~\cite{llrm}, as our semi-explicit representation already compresses scene information into a much smaller set of Gaussians, making Gaussian pruning unnecessary for training at our highest-resolution setup. We also remove the soft depth supervision from~\cite{llrm}, which is used to regularize 3D Gaussian positions. In \methodname{}, Gaussians are free to distribute anywhere in 3D space—rather than being constrained to object surfaces—as long as they enable accurate color and depth rendering at the target camera views.

\begin{figure*}
    \centering
    \includegraphics[width=0.9\linewidth]{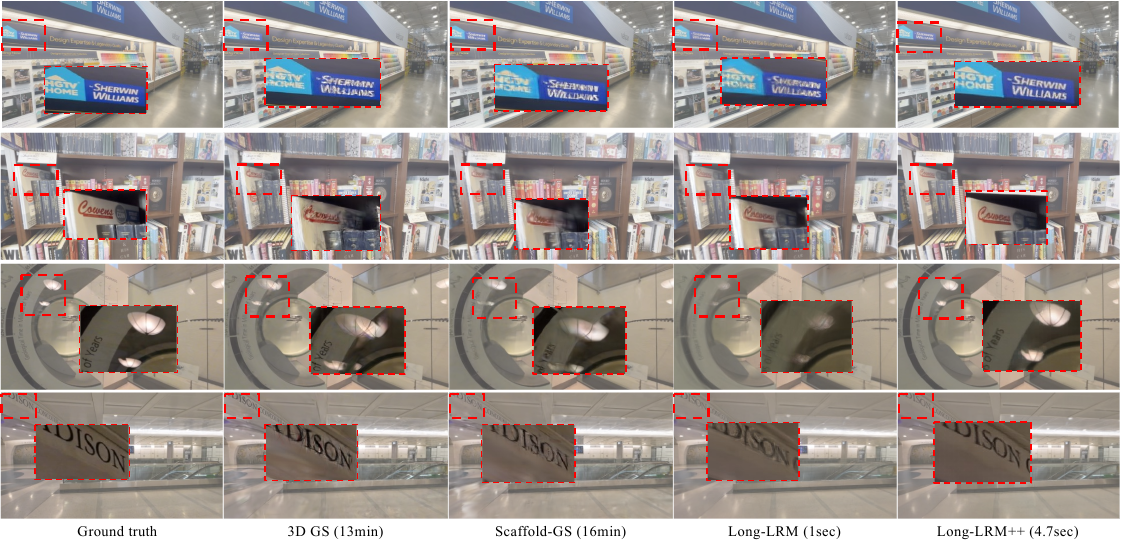}
    \vspace{-0.1in}
    \caption{\small Qualitative comparison on reconstruction details of DL3DV (32-input, 960×540-resolution) with baselines, including optimization-based methods. Reconstruction time in parentheses. \methodname{} notably produces fine details such as readable text in just 5 seconds.}
    \label{fig:text_comp}
\vspace{-0.1in}
\end{figure*}

\section{Experiments}
\label{sec:experiment}

We train and evaluate \methodname{} on two tasks: novel-view color rendering on the DL3DV~\cite{dl3dv} dataset, and novel-view color and depth rendering on ScanNetv2~\cite{scannet}. Common implementation details are provided in Sec.~\ref{sec:imp}, while the training stage configuration and evaluation results for the two tasks are presented in Secs.~\ref{sec:dl3dv} and~\ref{sec:scannet}, respectively.


\subsection{Implementation Details}\label{sec:imp}

Unless otherwise specified, we follow most training and architecture settings of~\cite{llrm}, including learning rate, optimizer, and data augmentation. The input processing backbone consists of 24 blocks arranged in a \{1T7M\}$\!\times\!3$ sequence with hidden dimensionality 768. The target-frame decoder comprises 5 transformer blocks in an `LLGGG' sequence (`L' for local attention, `G' for global attention) with hidden dimensionality 256. Following~\cite{lvsm}, we apply QK-Norm~\cite{qknorm} in all transformer blocks. Input images are patchified with patch size $p=8$. For each image token, we predict $K=16$ Gaussians with feature dimensionality $F=32$. Feature Gaussians are rendered onto a canvas of size $\frac{H}{r}\times \frac{W}{r}$ with $r=2$. Before the target-frame decoder, the rendered feature map is patchified into shape $\mathbb{R}^{\frac{H}{q}\times \frac{W}{q}\times E}$ with $q=8$ and $E=256$. For multi-space partitioning and merging, we set $S=2$.


\setlength{\tabcolsep}{1pt}
\begin{table}[h]
\centering

\begin{minipage}[t]{\linewidth}
\begin{center}
    \captionsetup{type=table}
    \hspace{-0.1in}\resizebox{\linewidth}{!}{
\begin{scriptsize}
\renewcommand{\arraystretch}{1.2}
\begin{tabular}{@{}$c@{\ \ }^l@{ }^c@{ }^c@{ }^c@{ }^c@{ }^c@{ }^c@{}}
\toprule
\rowstyle{\bfseries}
\makecell{Input\\Views} & \multirow{1}{*}{Method} &   PSNR$\uparrow$ & SSIM$\uparrow$ & LPIPS$\downarrow$ & \makecell{Reconstruction\\Time$\downarrow$} & \makecell{Rendering\\FPS$\uparrow$} \\
 \midrule
 \multirow{4}{*}{16} & 3D GS$_{30k}$ &  21.20 & 0.708 & 0.264
 &  13min & 50+\\
& Long-LRM & 22.66 & 0.740 & 0.292 &\cellcolor{first} 0.4sec & 50+ \\
& LaCT  & \cellcolor{first}24.70 & \cellcolor{second}0.793 &\cellcolor{first} 0.224 & 14.6sec & 1.8 \\
& Ours  & \cellcolor{second} 24.40 & \cellcolor{first}0.795 &\cellcolor{second} 0.231 &\cellcolor{second} 1.6sec & 14 \\
 \midrule
\multirow{6}{*}{32} & 3D GS$_{30k}$ &  23.60 & 0.779 & 0.213 &  13min & 50+ \\
& Scaffold-GS$_{30k}$ &  24.77  & 0.805 & 0.205 &  16min & 50+ \\
& Long-LRM  & 24.10 & 0.783 & 0.254 & \cellcolor{first}1sec & 50+ \\
& Long-LRM$_{10}$  &  25.60 & 0.826 & 0.233 & 37sec & 50+\\
& LaCT  & \cellcolor{first}26.90 & \cellcolor{second}0.837 & \cellcolor{second}0.185 & 29.3sec & 1.8\\
& Ours  & \cellcolor{second} 26.43 & \cellcolor{first}0.846 & \cellcolor{first}0.180 & \cellcolor{second}4.7sec & 14\\
\midrule
\multirow{4}{*}{64} & 3D GS$_{30k}$ & 26.43 & 0.854 & \cellcolor{second}0.167
& 14min & 50+\\
& Long-LRM  &   24.77 & 0.804 & 0.239 & \cellcolor{first}3sec & 50+ \\
& LaCT  &  \cellcolor{first}28.30 & \cellcolor{second}0.857 & 0.169 & 59sec & 1.8\\
& Ours  & \cellcolor{second} 27.30 & \cellcolor{first}0.869 &\cellcolor{first} 0.161 &\cellcolor{second} 16.3sec & 14 \\
\bottomrule
\end{tabular}
\end{scriptsize}}
\end{center}
\vspace{-0.6cm}
\caption{\textbf{Quantitative comparison of novel-view synthesis quality on the DL3DV-140 at 960×540 resolution.} Subscripts denote the number of (post-prediction) optimization steps. `Reconstruction Time' measures the latency of converting input images into the scene representation, while `Rendering FPS' reports the frame rate for generating RGB images from the representation. All timing results are measured on a single A100 GPU. \methodname{} is trained using 32 input views, and the results for 16- and 64-view settings are obtained in a zero-shot manner.
}
\label{tb:nvs}
\end{minipage}
\vspace{-0.2in}
\end{table}

\subsection{Novel-view color rendering on DL3DV}\label{sec:dl3dv}

\noindent\textbf{Dataset.}
DL3DV~\cite{dl3dv} is a large-scale, scene-level novel-view synthesis dataset containing diverse indoor and outdoor scenes. Camera poses are obtained via COLMAP~\cite{colmap}. We train on the DL3DV-10K split, which includes over 10K scenes (excluding those in DL3DV-140), and evaluate on the DL3DV-140 benchmark, which consists of 140 representative scenes. For each scene, we sample every 8th frame as targe views and sample input views from the remaining frames using K-means clustering as in Long-LRM.

\noindent \textbf{Training stages.}
Following~\cite{llrm}, we adopt a curriculum training schedule divided into four stages, gradually increasing the image resolution from $256\!\times\!256$ to $960\!\times\!540$. Unlike~\cite{llrm}, which starts directly with 32 input views, we begin with 8 input frames and scale to 32 only in the final stage to reduce the computation. See Sec.~\ref{sec:stage} for details.

\noindent \textbf{Evaluation results.}
Quantitative evaluation results are shown in Table~\ref{tb:nvs}. Our baselines include the optimization-based 3D GS~\cite{3dgs}, Long-LRM~\cite{llrm}, which predicts an explicit set of Gaussians, and LaCT~\cite{lact}, which stores scene information implicitly in the fast weights of TTT~\cite{ttt} layers. \methodname{} is trained only on the 32-view setup and evaluated zero-shot on 16- and 64-view settings, demonstrating strong generalization to different input lengths.
Compared to 3D GS and Long-LRM, \methodname{} achieves superior rendering quality across all input lengths and metrics. Notably, for 32 inputs, \methodname{} surpasses Long-LRM's result even after 10 post-prediction optimization steps.
For 64 inputs, Long-LRM underperforms the optimization-based 3D GS, whereas \methodname{} surpasses 3D GS by 0.9dB in PSNR. While  slower than color-based Gaussians (50+ FPS), \methodname{} still maintains real-time rendering at 14 FPS.
Compared to the fully implicit LaCT, \methodname{} achieves comparable rendering quality across all input sequences—especially excelling in SSIM—while rendering 8× faster, highlighting the benefits of our semi-explicit representation design.


\setlength{\tabcolsep}{1pt}
\begin{table}[h]
\centering

\begin{minipage}[t]{\linewidth}
\begin{center}
    \captionsetup{type=table}
    \hspace{-0.1in}\resizebox{\linewidth}{!}{
\begin{scriptsize}
\renewcommand{\arraystretch}{1.2}
\begin{tabular}{@{}$c@{\ \ }^l@{ }^c@{ }^c@{ }^c@{ }^c@{ }^c@{ }^c@{ }^c@{}}
\toprule
\rowstyle{\bfseries}
\multirow{2}{*}{\makecell{Input\\Views}} & \multirow{2}{*}{Method} &  \multicolumn{3}{^c}{Color rendering} & \multicolumn{4}{^c}{Depth rendering} \\
\cmidrule(lr){3-5}\cmidrule(lr){6-9}
&&   PSNR$\uparrow$ & SSIM$\uparrow$ & LPIPS$\downarrow$ &  Abs Diff$\downarrow$ & Abs Rel$\downarrow$ & Sq Rel$\downarrow$ & $\mathbf{\delta\!<\!1.25}$$\uparrow$ \\
 \midrule
 \multirow{2}{*}{128} & Long-LRM & 22.36 & 0.727 & 0.359 & 0.184 & 0.113 & 0.044 & 0.870  \\
& Ours & \cellcolor{first}26.96 & \cellcolor{first}0.807 &  \cellcolor{first}0.279 & \cellcolor{first}0.131 & \cellcolor{first}0.080 &\cellcolor{first} 0.023 & \cellcolor{first}0.935 \\
\bottomrule
\end{tabular}
\end{scriptsize}}
\end{center}
\vspace{-0.6cm}
\caption{\textbf{Quantitative comparison of novel-view color and depth rendering quality on the ScanNetv2 test split at 448×336 resolution.} For each scene, we sample every 8th frame from the first 1280 frames as target views, and uniformly sample 128 frames from the remaining sequence as input views.
}
\label{tb:depnvs}
\end{minipage}
\vspace{-0.2in}
\end{table}

\subsection{Novel-view rendering on ScanNetv2}\label{sec:scannet}

\noindent \textbf{Dataset.}
ScanNetv2~\cite{scannet} is a large-scale indoor dataset widely used for 3D reconstruction. Each RGB stream is paired with corresponding depth maps and ground-truth camera poses. The training split contains 1,201 rooms, while the test split includes 100 rooms.

\noindent \textbf{Training stages.}
We adopt a curriculum, multi-stage training strategy, gradually increasing the image resolution from $256\!\times\!256$ to $448\!\times\!336$ and the number of input frames from 8 to 128. See Sec.~\ref{sec:stage} for details.

\noindent \textbf{Evaluation results.}
Table~\ref{tb:depnvs} presents quantitative evaluation results comparing \methodname{} with Long-LRM for novel-view color and depth rendering on ScanNetv2. We train Long-LRM using the same training stages and training objectives as \methodname{}, directly supervising depth maps rendered from the 3D Gaussians. During evaluation, we use the first 1,280 frames of each scene, uniformly sampling every 8th frame as target views and 128 frames from the remaining frames as input.
\methodname{} demonstrates superior performance for both color and depth rendering. For color, it achieves a +4.6dB PSNR improvement, while for depth, it reduces the Absolute Difference metric by 0.053.

\begin{figure*}[h]
    \centering
    \includegraphics[width=0.8\linewidth]{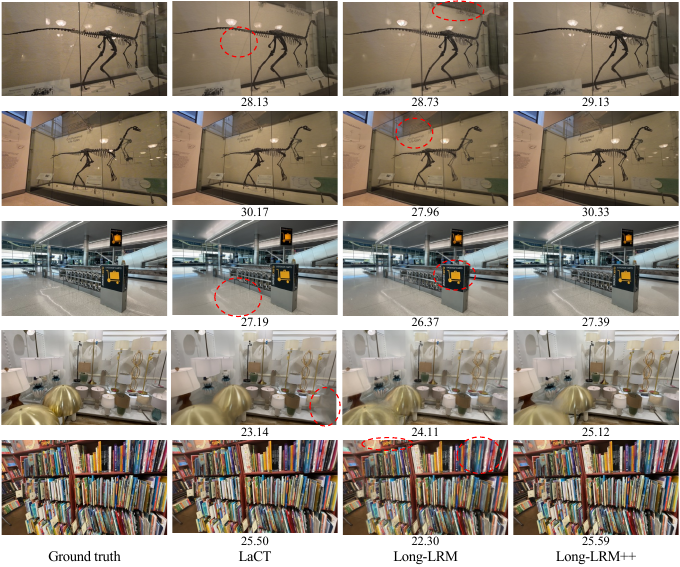}
    \vskip -0.05in
    \caption{\small 
    More qualitative comparison on DL3DV (32-input, 960×540-resolution). PSNR is shown below each rendered image. \methodname{} notably sharpens fine details compared to Long-LRM, and produces more faithful light reflections than LaCT.
}
    \label{fig:dl3dv}
    \vspace{-0.1in}
\end{figure*}

\begin{figure*}[h]
    \centering
    \includegraphics[width=0.8\linewidth]{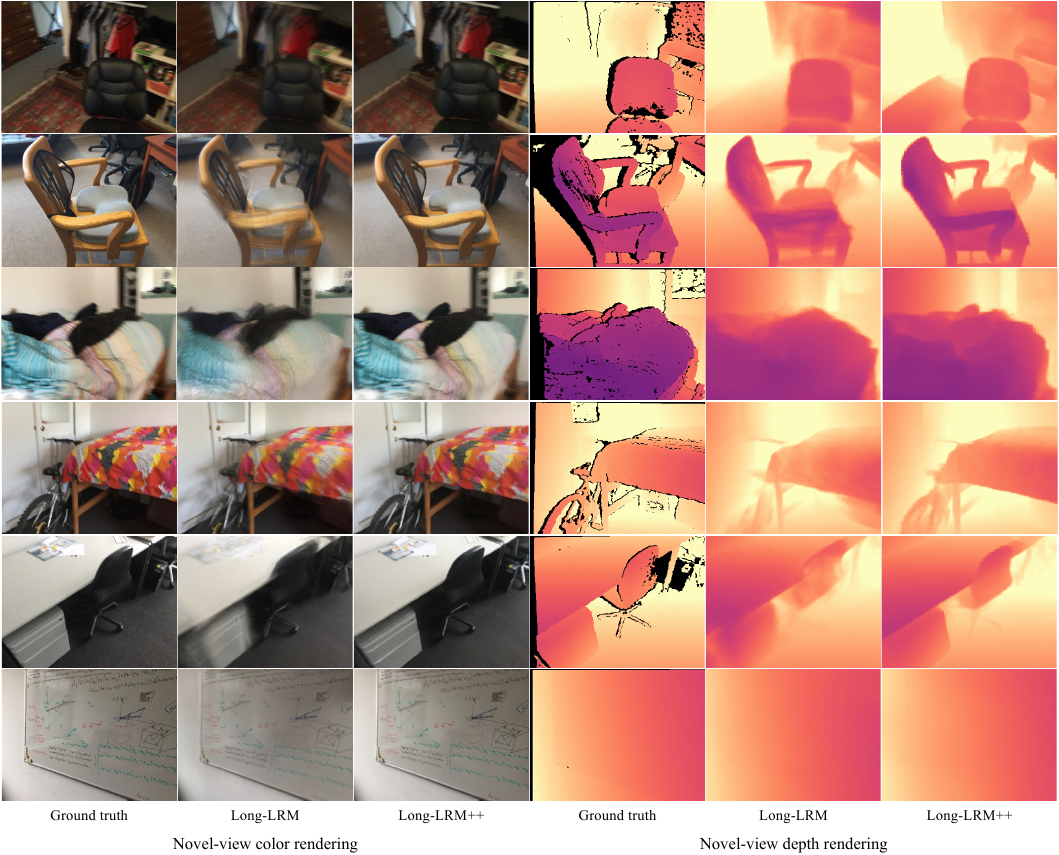}
    \vskip -0.05in
    \caption{\small Qualitative comparison of novel-view color and depth rendering on ScanNetv2 (128-input, 448×336 resolution).
    \methodname{} outperforms Long-LRM and produces high-quality depth maps despite using a sparse set of free-moving feature Gaussians.
}
    \label{fig:scannet}
    \vspace{-0.2in}
\end{figure*}

\section{Analysis}
\label{sec:analysis}


\setlength{\tabcolsep}{2pt}
\begin{table}[h]
\centering

\begin{minipage}[t]{0.52\linewidth}
\begin{center}
    \captionsetup{type=table}
    \hspace{-0.1in}\resizebox{\linewidth}{!}{%
\begin{scriptsize}
\renewcommand{\arraystretch}{1.2}
\begin{tabular}{@{}$c@{ }^c@{ }^c@{ }^c@{}}
\toprule
\rowstyle{\bfseries}
\makecell{\#Gaussian \\ per token} & \makecell{Gaussian \\feature dim} & \makecell{Feature splat\\canvas size} & PSNR$\uparrow$ \\
\midrule
8 & 32 & 0.5 & 27.94 \\
8 & 64 & 0.5 &  28.20 \\
\hdashline
16 & 32 & 0.25 & 27.81 \\
\rowcolor{chosen} 16 & 32 & 0.5 & 28.37 \\
16 & 32 & 1.0 & 28.41 \\
\hdashline
16 & 64 & 0.5 & 28.44  \\
32 & 16 & 0.5 & 26.71\\
\bottomrule
\end{tabular}
\end{scriptsize}}
\end{center}
\vspace{-0.6cm}
\caption{Ablation studies on number of Gaussians per token, Gaussian feature dimensionality, and feature-splat canvas size (ratio to input size), evaluated on DL3DV-140 under the 4-view, 256-resolution setting with the ``LLG" decoder. \colorbox{chosen}{row} is the chosen setup.
}
\label{tb:featsplat}
\end{minipage}
\hfill
\begin{minipage}[t]{0.46\linewidth}
\begin{center}
    \captionsetup{type=table}
    \hspace{-0.1in}\resizebox{\linewidth}{!}{%
\begin{scriptsize}
\renewcommand{\arraystretch}{1.2}
\begin{tabular}{@{}$l@{ }^c@{ }^c@{ }^c@{}}
\toprule
\rowstyle{\bfseries}
\makecell{Splat\\type} & \makecell{Decoder\\blocks} & \makecell{Decoder\\dim} & PSNR$\uparrow$ \\
\midrule
Color & / & / & 27.32\\
Feature & GGG & 256 & 27.78 \\ 
Feature & LLL & 256 & 28.23 \\ 
Feature & LLG & 256 & 28.37\\ 
\rowcolor{chosen} Feature & LLGGG & 256 & 28.74 \\ 
Feature & LLGGG & 768 & 29.09 \\ 
\bottomrule
\end{tabular}
\end{scriptsize}}
\end{center}
\vspace{-0.6cm}
\caption{Ablation studies on decoder block architecture and dimensionality. ``L" denotes a local-attention layer and ``G" denotes a global-attention layer, evaluated on DL3DV-140 under the 4-view, 256-resolution setting.
}
\label{tb:decoder}
\end{minipage}
\vspace{-0.1in}
\end{table}

\setlength{\tabcolsep}{1pt}
\begin{table}[h]
\centering
\begin{minipage}[t]{0.95\linewidth}
\begin{center}
    \captionsetup{type=table}
    \hspace{-0.1in}\resizebox{\linewidth}{!}{%
\begin{scriptsize}
\renewcommand{\arraystretch}{1.2}
\begin{tabular}{@{}$c@{ }^c@{ }^c@{ }^c@{ }^c@{ }^c@{ }^c@{ }^c@{ }^c@{}}
\toprule
\rowstyle{\bfseries}
\multirow{2}{*}{\makecell{\#Partition}} & \multicolumn{3}{^c}{8-input,256-res} & \multicolumn{5}{^c}{32-view, 960$\!\times\!$540}\\
\cmidrule(lr){2-4}\cmidrule(lr){5-9}
& PSNR$\uparrow$ & SSIM$\uparrow$ & LPIPS$\downarrow$ & PSNR$\uparrow$ & SSIM$\uparrow$ & LPIPS$\downarrow$ & \makecell{Recon Time$\downarrow$} & \makecell{Render FPS$\uparrow$} \\
\midrule
1 & 28.74 & 0.906 & 0.093 & 26.08 & 0.836 & 0.191 & 4.7sec & 18 \\
\rowcolor{chosen}2 & 29.73 & 0.919 & 0.084 & 26.43 & 0.846 & 0.180 &  4.7sec & 14 \\
3 & 29.82 & 0.922 & 0.074 &  - &  - &  - &  - &  - \\
\bottomrule
\end{tabular}
\end{scriptsize}}
\end{center}
\vspace{-0.6cm}
\caption{Ablation studies on the number of Gaussian partitions, evaluated on DL3DV-140.
}
\label{tb:featsplit}
\end{minipage}
\vspace{-0.1in}
\end{table}


\begin{table}[h]
\centering

\begin{minipage}[t]{0.8\linewidth}
\begin{center}
    \captionsetup{type=table}
    \hspace{-0.1in}\resizebox{\linewidth}{!}{
\begin{scriptsize}
\renewcommand{\arraystretch}{1.2}
\begin{tabular}{@{}$l@{ }^c@{ }^c@{ }^c@{ }^c@{ }^c@{}}
\toprule
\rowstyle{\bfseries}
 \multirow{2}{*}{Method} &  \multicolumn{3}{^c}{Color rendering} & \multicolumn{2}{^c}{Depth rendering} \\
\cmidrule(lr){2-4}\cmidrule(lr){5-6}
 &PSNR$\uparrow$ & SSIM$\uparrow$ & LPIPS$\downarrow$ &  Abs Diff$\downarrow$ & $\mathbf{\delta\!<\!1.25}$$\uparrow$ \\
 \midrule
 Long-LRM$_{\text{2DGS}}$ & 23.89 & 0.744 & 0.344 & 0.227 & 0.821 \\
 Long-LRM$_{\text{3DGS}}$ & 24.30 & 0.759 & 0.321 & 0.243 & 0.818 \\
 Ours$_{\text{w/o aux losses}}$ & \cellcolor{first} 28.28 & \cellcolor{first} 0.835 & \cellcolor{first} 0.226 & 0.157 & 0.911 \\
 Ours & 27.86 & 0.826 & 0.234 & \cellcolor{first}0.135 & \cellcolor{first} 0.916 \\
\bottomrule
\end{tabular}
\end{scriptsize}}
\end{center}
\vspace{-0.6cm}
\caption{Ablation studies on novel-view color and depth rendering on ScanNetv2 under the 8-input, 256-resolution setup. Methods are trained for 10K steps.
}
\label{tb:dep_ab}
\end{minipage}
\vspace{-0.2in}
\end{table}

\subsection{Hyperparameter studies for feature rendering}

In Table~\ref{tb:featsplat}, we present the rendering quality under different hyperparameter configurations for the feature Gaussian design, including the number of Gaussians $K$ predicted by each image token, the feature vector dimensionality $F$ of each Gaussian, and the size of the feature rendering canvas.
Intuitively, increasing either $K$ or $F$ allows the model to store more information about the scene, thereby improving rendering quality. A larger $K$ enables finer spatial resolution and more flexible distribution of features in 3D space, whereas a larger $F$ allows each Gaussian to encode more complex local information, and also increases the “thickness” of the rendered feature map.
However, indiscriminately increasing $K$ or $F$ can result in substantially higher memory consumption. Therefore, we need to strike a balance between the two. To investigate this trade-off, we compare three configurations: `$K8,F64$', `$K16,F32$', `$K32,F16$'. These settings have roughly similar memory footprint, while `$K16,F32$' achieves the best rendering quality, suggesting that this configuration provides the optimal balance between spatial flexibility (controlled by $K$) and feature complexity (controlled by $F$).
The rendering canvas size determines the resolution of the rendered feature map. Increasing it generally improves rendering quality, but the benefit diminishes as the resolution grows. Taking both rendering quality and efficiency into account, we select the \colorbox{chosen}{highlighted row} as our final configuration. Note that $F=32$ is smaller than the 48 channels used by color Gaussians with SH degree 3.

\subsection{Ablation studies for feature decoding}

\noindent\textbf{Target-frame decoder.} Table~\ref{tb:decoder} presents the rendering quality under different decoder block configurations and hidden dimensionalities. Comparing `GGG' and `LLL', we observe that using purely local-attention blocks yields better performance than purely global-attention ones. Further, the hybrid `LLG', where local-attention blocks are followed by global attention, achieves even higher quality, suggesting that global feature aggregation is beneficial. As expected, increasing the hidden dimensionality from 256 to 768 also improves performance. Balancing rendering quality and efficiency, we adopt the \colorbox{chosen}{highlighted row} as our final configuration.

\noindent\textbf{Multi-space partitioning and merging.} Table~\ref{tb:featsplit} demonstrates the effectiveness of the proposed partitioning and merging design. Increasing the partition count $S$ from 1 to 2 consistently improves performance across all three rendering quality metrics, while only slightly affecting rendering speed. However, further increasing $S$ to 3 yields diminishing returns in performance under the low-resolution setup, so we opt not to adopt it in our final configuration.

\subsection{Ablation studies for novel-view depth rendering}

Table~\ref{tb:dep_ab} presents ablation studies for the novel-view depth rendering task. We compare the performance of Long-LRM with a 3D Gaussian head, Long-LRM with a 2D Gaussian head~\cite{2dgs}, and \methodname{}. The results show that Long-LRM$_{\text{3DGS}}$ achieves higher color rendering quality but slightly worse depth accuracy than Long-LRM$_{\text{2DGS}}$, while \methodname{} outperforms both by a substantial margin on both color and depth rendering metrics.
We further ablate the two auxiliary losses used for depth supervision—the gradient loss and the normal loss. The results indicate that these auxiliary signals effectively enhance depth rendering quality, though they slightly degrade color performance. We speculate this trade-off arises because these losses encourage the decoder to allocate more capacity toward depth reconstruction rather than color decoding.
\section{Conclusion}
\label{sec:conclusion}

We introduced \methodname{}, a feed-forward novel-view synthesis framework for high-resolution scene reconstruction. In contrast to Long-LRM, which predicts pixel-aligned color Gaussians, \methodname{} leverages a set of free-moving feature Gaussians that provide greater robustness and stronger representational capacity. Together with a lightweight decoder that incorporates translation-invariant local-attention blocks, as well as a novel multi-space partitioning–and–merging mechanism, \methodname{} effectively resolves the blurriness artifacts observed in Long-LRM and delivers a substantial boost in rendering quality. Our method achieves performance competitive with fully implicit approaches such as LaCT, yet maintains real-time rendering speed, whereas LaCT suffers from slow inference. Furthermore, the proposed semi-explicit representation reduces the number of required Gaussians, lowers storage cost, and improves robustness for long input sequences. Overall, \methodname{} demonstrates an efficient and scalable path toward high-quality, high-resolution feed-forward 3D scene reconstruction.
{
    \small
    \bibliographystyle{ieeenat_fullname}
    \bibliography{main}
}

\clearpage
\setcounter{page}{1}
\maketitlesupplementary

\section{More implementation details}

Due to its semi-explicit formulation, \methodname{} exhibits a stronger tendency to overfit to input frames when training on mixed sets of input and unseen target frames. This effect becomes more pronounced on datasets such as DL3DV, where neighboring frames have relatively large pose differences—that is, the effective frame density per unit scene coverage is lower. To mitigate this overfitting, we manually reduce the probability that input frames are selected as target frames during training. Concretely, during the random sampling (without replacement) of target frame indices, we decrease the selection weight of input frames to 0.1 while keeping all other frames at 1. For denser datasets such as ScanNetv2, this adjustment is unnecessary because the number of unseen frames significantly exceeds the number of input frames.

\begin{figure}
    \centering
    \begin{minipage}[t]{0.43\linewidth}
    \includegraphics[width=0.99\linewidth]{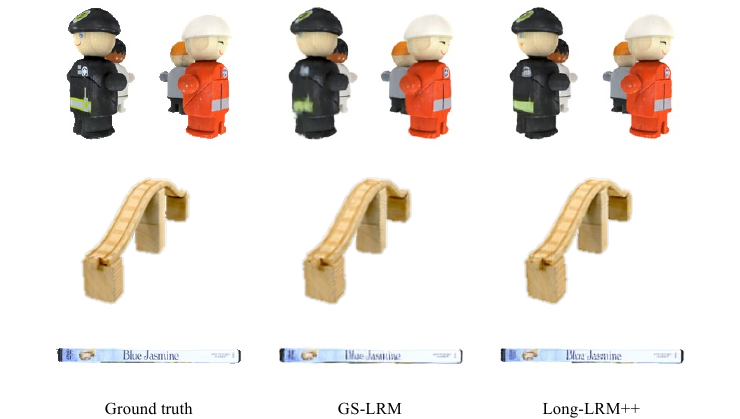}
    \vspace{-0.2in}
    \caption{\footnotesize Qualitative comparison with GS-LRM for object reconstruction on the GSO dataset.}
    \label{fig:obj}
    \end{minipage}
    \hfill
    \begin{minipage}[t]{0.54\linewidth}
    \includegraphics[width=0.99\linewidth]{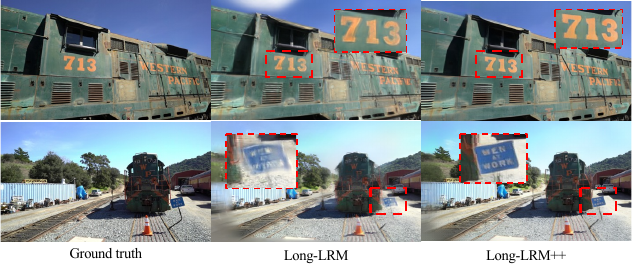}
    \vspace{-0.2in}
    \caption{\footnotesize Qualitative comparison with Long-LRM on the Tanks\&Temples dataset.}
    \label{fig:tnt}
    \end{minipage}
\end{figure}

\setlength{\tabcolsep}{0pt}
\begin{table}[h]
\centering
\begin{minipage}[t]{0.32\linewidth}
\begin{center}
    \captionsetup{type=table}
    \hspace{-0.1in}\resizebox{\linewidth}{!}{
\begin{scriptsize}
\begin{tabular}{@{}$l@{ }^c@{}}
\toprule
\rowstyle{\bfseries}
 Method & PSNR$\uparrow$\\
 \midrule
 GS-LRM$_{\text{dim768}}$ & 31.39 \\
GS-LRM & 31.95 \\
\methodname{} & \cellcolor{first}32.52\\
\bottomrule
\end{tabular}
\end{scriptsize}}
\end{center}\vspace{-0.6cm}
\caption{\footnotesize Object reconstruction on GSO.
}
\label{tb:gso}
\end{minipage}
\hfill
\begin{minipage}[t]{0.32\linewidth}
\begin{center}
    \captionsetup{type=table}
    \hspace{-0.1in}\resizebox{\linewidth}{!}{
\begin{scriptsize}
\begin{tabular}{@{}$l@{ }^c@{}}
\toprule
\rowstyle{\bfseries}
 Method & PSNR$\uparrow$\\
 \midrule
GS-LRM & 28.10 \\
Long-LRM & 28.54 \\
\methodname{} & \cellcolor{first}29.31\\
\bottomrule
\end{tabular}
\end{scriptsize}}
\end{center}\vspace{-0.6cm}
\caption{\footnotesize Scene reconstruction on RE10K.
}
\label{tb:re10k}
\end{minipage}
\hfill
\begin{minipage}[t]{0.32\linewidth}
\begin{center}
    \captionsetup{type=table}
    \hspace{-0.1in}\resizebox{\linewidth}{!}{
\begin{scriptsize}
\begin{tabular}{@{}$l@{ }^c@{}}
\toprule
\rowstyle{\bfseries}
 Method & PSNR$\uparrow$\\
 \midrule
3D GS & 18.10 \\
Long-LRM & 18.38 \\
\methodname{} & \cellcolor{first}19.30\\
\bottomrule
\end{tabular}
\end{scriptsize}}
\end{center}\vspace{-0.6cm}
\caption{\footnotesize Scene reconstruction on T\&T.
}
\label{tb:tnt}
\end{minipage}
\vspace{-0.1in}
\end{table}

\section{More evaluation results}

We evaluate \methodname{} on three additional datasets: object reconstruction on GSO (Table~\ref{tb:gso}), scene reconstruction on RealEstate10K (Table~\ref{tb:re10k}), and zero-shot scene reconstruction on Tanks\&Temples (Table~\ref{tb:tnt}). For GSO, we train \methodname{} on Objaverse for 80K steps and evaluate on GSO, comparing against GS-LRM and GS-LRM$_{\text{dim768}}$, which matches \methodname{}’s backbone dimension. As shown in Fig.~\ref{fig:obj}, \methodname{} achieves superior detail fidelity. For RealEstate10K, we train on the training split for 100K steps and evaluate on the test split, achieving state-of-the-art performance. For Tanks\&Temples, we conduct zero-shot evaluation using a model trained on DL3DV, obtaining a +1 dB PSNR improvement over Long-LRM (see qualitative results in Fig.~\ref{fig:tnt}).

\section{Comparison with Depth Anything 3 (DA3)}

\begin{figure*}
    \centering
    \includegraphics[width=0.9\linewidth]{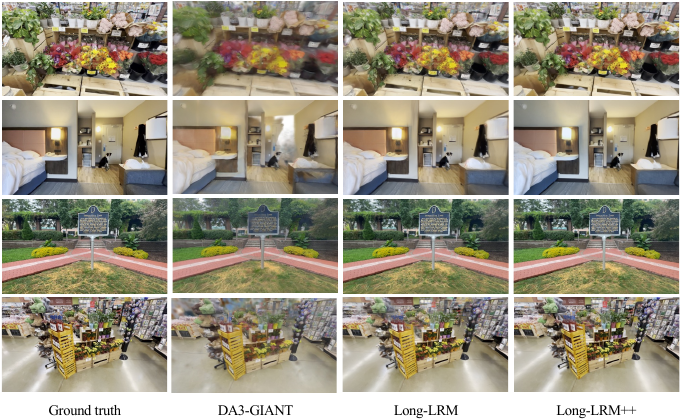}
    \vspace{-0.1in}
    \caption{\small Qualitative comparison with DA3 on DL3DV (32-input, 960×540-resolution) using DA3 poses.}
    \label{fig:da3}
\vspace{-0.1in}
\end{figure*}

We compare \methodname{} (110M param) with the Gaussian prediction feature of DA3-GIANT~\cite{da3} (1.15B param) on DL3DV using both COLMAP poses and DA3 poses. To obtain DA3 poses, we run the pose predictor of DA3 on all frames of a scene. During inference, we feed the obtained poses of selected input frames to the Gaussian prediction models. Quantitative comparison is shown in Table~\ref{tb:da3} and qualitative in Fig.~\ref{fig:da3}. Both Long-LRM and \methodname{} show better rendering quality than DA3-GIANT.

\setlength{\tabcolsep}{1pt}
\begin{table}[h]
\centering

\begin{minipage}[t]{0.7\linewidth}
\begin{center}
    \captionsetup{type=table}
    \hspace{-0.1in}\resizebox{\linewidth}{!}{
\begin{scriptsize}
\renewcommand{\arraystretch}{1.2}
\begin{tabular}{@{}$c@{\ \ }^l@{ }^c@{ }^c@{ }^c@{ }^c@{ }^c@{ }^c@{}}
\toprule
\rowstyle{\bfseries}
\makecell{Pose Source} & \multirow{1}{*}{Method} &   PSNR$\uparrow$ & SSIM$\uparrow$ & LPIPS$\downarrow$ \\
 \midrule
 \multirow{3}{*}{COLMAP} & DA3-GIANT &  17.52 & 0.562 & 0.382 \\
& Long-LRM & 24.10 & 0.783 & 0.254  \\
& \methodname{}  & \cellcolor{first} 26.43 & \cellcolor{first}0.846 & \cellcolor{first}0.180 \\
 \midrule
\multirow{3}{*}{DA3} & DA3-GIANT &  17.27 & 0.540 & 0.395 \\
& Long-LRM  & 22.98 & 0.731 & 0.277 \\
& \methodname{}  & \cellcolor{first} 24.43 & \cellcolor{first}0.773 & \cellcolor{first}0.212 \\
\bottomrule
\end{tabular}
\end{scriptsize}}
\end{center}
\vspace{-0.6cm}
\caption{Quantitative comparison with DA3 on DL3DV-140 (32 input views, 960×540 resolution).
}
\label{tb:da3}
\end{minipage}
\vspace{-0.2in}
\end{table}

\section{Training stage configuration}\label{sec:stage}

Table~\ref{tb:stage} summarizes the detailed setup for each stage of training of \methodname{} on DL3DV, including training iterations, number of GPUs, and total GPU hours. Table~\ref{tb:stage_dep} summarizes the stage setup for the ScanNetv2 training.

\begin{table}[h]
\centering

\begin{minipage}[t]{\linewidth}
\begin{center}
    \captionsetup{type=table}
    \hspace{-0.1in}\resizebox{\linewidth}{!}{
\begin{scriptsize}
\renewcommand{\arraystretch}{1.2}
\begin{tabular}{@{}$c@{ }^c@{ }^c@{ }^c@{ }^c@{ }^c@{ }^c@{ }^c@{ }^c@{}}
\toprule
\rowstyle{\bfseries}
\makecell{Stage} & \#Input &  \#Target & Resolution & Time/Step & \#Step & \makecell{Batch\\size} & \#GPU & \makecell{GPU\\Hours} \\
\midrule
1 & 8 & 8 & 256$\!\times\!$256 & 9.6sec & 60K & 256 & 16 & 2560 \\
2 & 8 & 8 & 512$\!\times\!$512 & 7.7sec & 10K & 64 & 16 & 342 \\
3 & 8 & 8 & 960$\!\times\!$540 &  17.6sec & 10K & 64 & 16 & 782 \\
4 & 32 & 8 & 960$\!\times\!$540 & 27.4sec & 10K & 64 & 64 & 4871 \\
\bottomrule
\end{tabular}
\end{scriptsize}}
\end{center}
\vspace{-0.6cm}
\caption{
Training stage configuration of \methodname{} for the DL3DV10K novel-view synthesis task. \textbf{GPU Hours} is calculated as \textbf{Time/Step $\times$ \#Steps $\times$ \#GPU}.
}
\label{tb:stage}
\end{minipage}
\vspace{-0.2in}
\end{table}

\begin{table}[h]
\centering

\begin{minipage}[t]{\linewidth}
\begin{center}
    \captionsetup{type=table}
    \hspace{-0.1in}\resizebox{\linewidth}{!}{
\begin{scriptsize}
\renewcommand{\arraystretch}{1.2}
\begin{tabular}{@{}$c@{ }^c@{ }^c@{ }^c@{ }^c@{ }^c@{ }^c@{ }^c@{ }^c@{}}
\toprule
\rowstyle{\bfseries}
\makecell{Stage} & \#Input &  \#Target & Resolution & Time/Step & \#Step & \makecell{Batch\\size} & \#GPU & \makecell{GPU\\Hours} \\
\midrule
1 & 8 & 8 & 256$\!\times\!$256 & 6.7sec & 20K & 128 & 8 & 298 \\
2 & 8 & 8 & 448$\!\times\!$336 & 3.9sec & 5K & 128 & 32 & 173 \\
3 & 128 & 8 & 448$\!\times\!$336 & 30.6sec & 2K & 64 & 64 & 1088 \\
\bottomrule
\end{tabular}
\end{scriptsize}}
\end{center}
\vspace{-0.6cm}
\caption{
Training stage configuration of \methodname{} for the ScanNetv2 novel-view color+depth rendering task. 
}
\label{tb:stage_dep}
\end{minipage}
\vspace{-0.2in}
\end{table}





\end{document}